# Delegating Custom Object Detection Tasks to a Universal Classification System

**Andrew Gleibman**

Sampletalk Research, POB 7141, Yokneam-Illit 20692, Israel
www.sampletalk.com

In this paper, the concept of multipurpose object detection system, recently introduced in [1], is discussed. Fig.1 below illustrates the innovative and the business aspects of this method: *transform a classifier into an object detector/locator via an image grid*. Classification techniques allow the application of powerful machine learning methods detailed below. A custom object detection system only needs to analyze the local classification results on the image grid. In this way, creation of a custom system is facilitated by first applying the universal approach and then doing a custom analysis of its results.

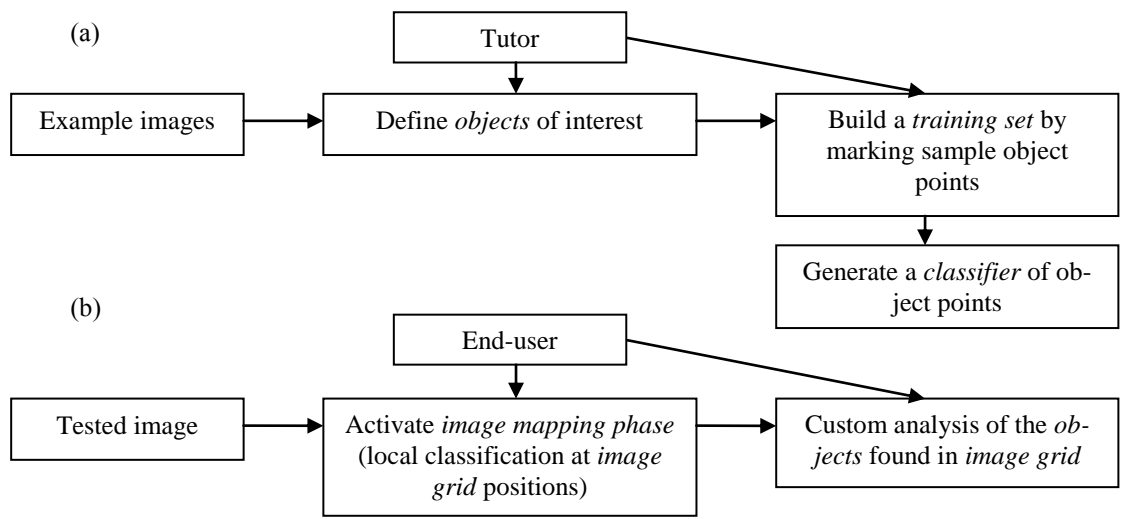

Fig.1. Delegating object detection task to a classifier: a) training phase, b) classification results in image grid positions are applied for custom reasoning about presence, absence, location, combination and behavior of the objects of interest.

The task of the *universal* system is creation of a classifier, which will supply the grid positions of the analyzed image with probabilities of belonging to the defined object classes.

The task of the *custom* component is standardized by considering only the above info at grid positions. For instance (Fig. 2), the detected points of class Pupil in grid positions on the frames of a video sequence are interpolated in order to rectify the position and reconstruct the motion of the human pupil.

This is essentially a universal framework for *locating objects of interest through classification*. An image may contain many objects of interest, so we apply the classifier locally in all image positions, belonging to the grid, assuming some constant-radius context regions around the analyzed positions. The user defines the grid resolution and the radius of the context regions.

The universality has huge advantages provided by modern machine learning methods. As an example, the pupil in Fig. 2 is successfully located in spite of complex natural-light illumination with light reflections and blurring, without any polarizing or filtering devices. No a priori info, no geometric models of the eye, no special calculations are necessary. Everything is done automatically by applying the *universal* techniques to raw training examples. As the universal techniques, in our experiments we apply SVM with RBF kernel [2], mRMR [3], and numerical feature algorithms described below.

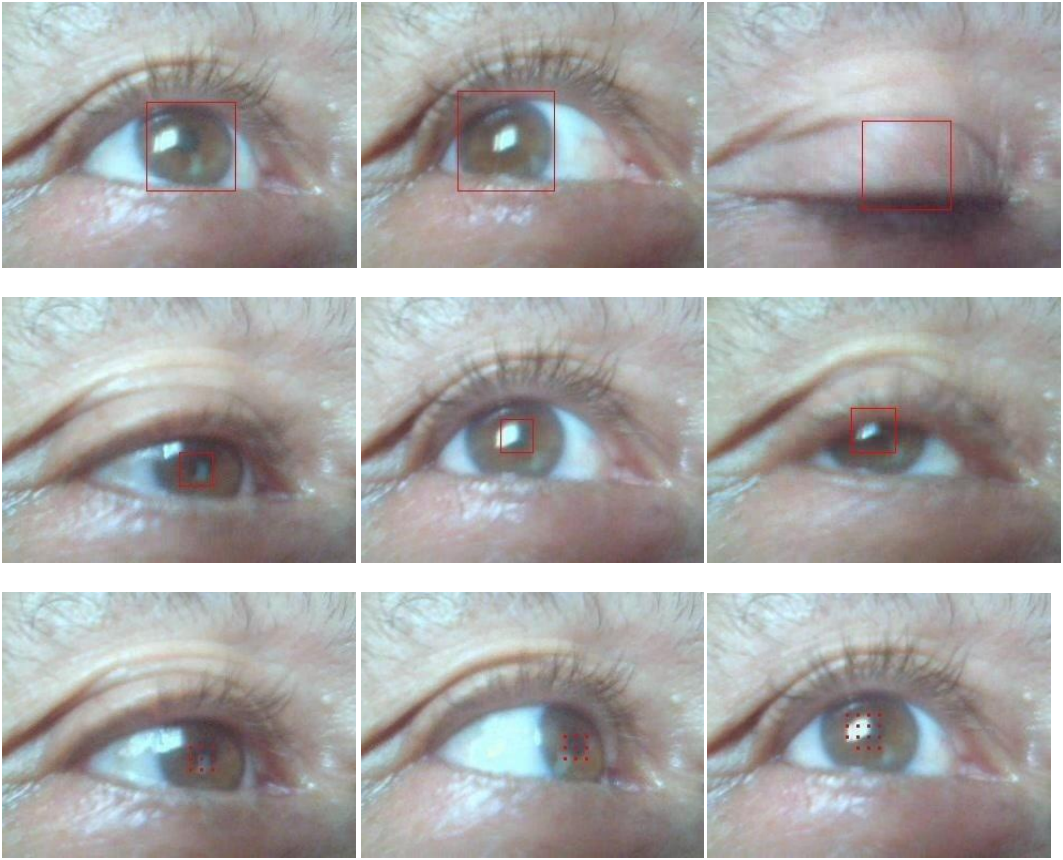

Fig.2. Detection of human iris (*top row*) and pupil (*middle row*) in complex natural illumination. When the eyelid is completely or partly closed (*right images of two upper rows*), the iris (pupil) position is found by an interpolation using consecutive frames of the video sequence. The *bottom row* shows actually detected points of class Pupil, which are applied for constructing the above enclosing rectangles and for the interpolation. See http://www.youtube.com/watch?v=QMlSgYYcOac for the live demo.

The pattern recognition component of such system should be as universal as possible for a wide class of applications related to computer vision. This approach assumes using the most general numerical features for classification, suitable for possible applications in a wide range of areas where the feature combinations are suitable for definition of classes in more specific areas. Among the applied methods we should mention those based on texture analysis and Fourier, Gabor and Wavelet transforms, with calculation of local power spectra in specific frequency bands. The custom component organizes these algorithms for performing so various tasks as raising alert messages, locating and counting special objects presented in the scanned scenes, detecting the presence, absence, or specific combination of objects, evaluating the dynamics of thereof etc. More details are given in [1].

The architecture and the numerical feature algorithms of the presented system are influenced by our research presented in [4]. An *object* can represent a complex and synthetic concept, which may include essentially different sub-classes. As an example, a concept of building construction can be exemplified by marking points of roof elements, wall fragments, chimneys of different kinds, windows, fences, wall joints etc. Paper [1] contains more examples of this kind.

Definition and analysis of so complex objects seems to be problematic for some popular frameworks such as the Viola-Jones's framework [5]. In our framework, the results of *image mapping phase* (cf. Fig.1) can be analyzed for various purposes. Viola-Jones's framework is more similar to a black box, which does not allow the user to analyze the plausibility assessment of object detection in individual image points. While we are concentrating on object classification and detection at exact image grid positions, the Viola-Jones's framework works with entire image regions. Additionally, the training time in Viola-Jones's algorithm is typically much larger than the training time in our case [1].

Additional innovative elements of this project are related to numerical features for classification and to automatic synthesis of the unlimited number of thereof.

In addition to known texture measurements based on co-occurrence matrices [6], we introduce a special *Orthogonal Gradient Co-occurrence Matrix (OGCM)*. For every pixel of the analyzed area, two values $g1, g2$ of image gradients in orthogonal directions are calculated. This can be done either for a (SN, EW) pair or for more pairs of orthogonal directions. The *OGCM* matrix is then built by incrementing element ($|g1|, |g2|$) of the matrix for every analyzed pixel. In a way, this matrix characterizes a 3D form of the objects presented in a 2D form. Our experiments show that when SVM is applied with RBF kernel, numerical feature values of homogeneity, entropy, contrast etc., typically extracted from co-occurrence matrices, can be avoided without any loss of the system functionality. Instead, the matrix elements themselves are applied as the numerical feature values for classification.

Classification, based on texture and power spectrum-related feature values as described above, is typically vulnerable to changes in image resolution. For a more robust presentation of image topology we introduce a special *Gray Level-Radius Co-occurrence Matrix (GLRCM)*. In order to build this matrix, a region around the tested image point is considered a set of concentric rings with external radii $w, 2w, 3w, …, Nw=R$, where $w$ is a constant width for all rings and $R$ is the radius of the context region. For every ring, a gray level histogram is calculated, where $G$ is the maximal gray level value. Then, the matrix of $N*G$ size is built by setting value of the ($n, g$) matrix element to the number of pixels of gray level $g$ in ring number $n$.

In most our experiments, the classification feature values taken immediately from this matrix lead to better results than those based on other texture methods. Indeed, *GLRCM* contains information about local image texture in combination with local shape characteristics independently on the image resolution. The important feature of *GLRCM* and *OGCM* is low calculation complexity, which is linear in the number of pixels in the analyzed region. This is essential when a special hardware implementation is assumed.

Numerical feature extraction methods, described above, are essentially parameterized by the following parameters: number $G$ of gray levels, ring width $w$, radius $R$ of the context regions. These parameters affect the dimension of texture representation matrices and the number of accumulated values in the matrix elements. Variation of these parameters substantially affects the number of numerical features, applied in classification algorithms, independently on feature selection methods. Additional parameters are mRMR-related feature selection thresholds for feature correlation and for mutual information with the defined classes [3].

In our experiments we apply a special system architecture, which allows variation of these parameters. This leads to *a controllable synthesis of unrestricted number of numerical features for classification*. The number of so generated numerical features in our experiments varies from a few dozen to a few thousand.

The proposed system architecture has many application areas and can be an important part for manufacturing specialized devices and systems. Our experiments show the applicability of this architecture to such areas as intrusion detection, pedestrian & motorcycle detection, general vehicle detection, lane departure & forward collision warning in transportation, eye fatigue detection, gaze tracking in complex illumination, special instrument detection in process control, pre-contact event detection, motion analysis etc.

## References

1. Gleibman, A. *Object Recognition System Design in Computer Vision: a Universal Approach*. Cornell University Library, Computer Vision and Pattern Recognition (cs.CV), October 27, 2013, http://arxiv.org/abs/1310.7170

2. Vapnik, V. N. *The Nature of Statistical Learning Theory* (2nd Ed.), Springer Verlag, 2000.

3. Peng, H.C., Long, F., and Ding, C. *Feature selection based on mutual information: criteria of max-dependency, max-relevance, and min-redundancy*. IEEE Trans. Patt. Analysis and Machine Intelligence, Vol. 27, No. 8, pp. 1226–1238, 2005.

4. Gleibman, A. H. *Intelligent Processing of an Unrestricted Text in First Order String Calculus*. Trans. on Comput. Sci. V, Special Issue on Cognitive Knowledge Representation, LNCS 5540, 2009, pp. 99–127. http://www.sampletalk.com/SampletalkLanguage/55400099.pdf

5. P. Viola, M. Jones. *Rapid Object Detection using a Boosted Cascade of Simple Features*. TR2004-043.

6. R.M. Haralick, K. Shanmugam, O. Dinstain. *Textural Features for Image Classification*. IEEE TRANS. ON SYS., MAN AND CYBRNETICS, V.SMC-3, No. 6, 1973, pp. 610-621.